\documentclass[10pt,twocolumn,letterpaper]{article}

\usepackage[pagenumbers]{cvpr}  % To force page numbers, e.g. for an arXiv version

\usepackage{capt-of}
\usepackage{xspace}
\usepackage{xcolor}
\usepackage{enumitem}
\usepackage{amsmath,amsthm,amssymb,amsfonts,dsfont,pifont,bm,bbm,mathrsfs,mathtools,nicefrac,extarrows,relsize}
\usepackage{algorithm,algpseudocode,listings}
\usepackage{booktabs,multirow,adjustbox,diagbox,threeparttable,tabularray,setspace}

\definecolor{cvprblue}{rgb}{0.21,0.49,0.74}
\usepackage[pagebackref,breaklinks,colorlinks,citecolor=cvprblue,bookmarks=false]{hyperref}
\usepackage{wrapfig}
\usepackage[capitalize]{cleveref}  % Should be loaded after 'hyperref', and works perfectly with 'subfigure'.

\crefname{section}{Sec.}{Secs.}
\Crefname{section}{Section}{Sections}
\crefname{appendix}{App.}{Apps.}
\Crefname{appendix}{Appendix}{Appendices}
\crefname{table}{Tab.}{Tabs.}
\Crefname{table}{Table}{Tables}
\crefname{figure}{Fig.}{Figs.}
\Crefname{figure}{Figure}{Figures}
\crefname{equation}{Eq.}{Eqs.}
\Crefname{equation}{Equation}{Equations}
\crefname{theorem}{Thm.}{Thms.}
\Crefname{theorem}{Theorem}{Theorems}
\crefname{lemma}{Lem.}{Lems.}
\Crefname{lemma}{Lemma}{Lemmas}
\crefname{remark}{Rem.}{Rems.}
\Crefname{remark}{Remark}{Remarks}
\crefname{corollary}{Cor.}{Cors.}
\Crefname{corollary}{Corollary}{Corollaries}
\crefname{algorithm}{Alg.}{Algs.}
\Crefname{algorithm}{Algorithm}{Algorithms}
\definecolor{cellred}{RGB}{213, 123, 101}
\definecolor{cellgreen}{RGB}{0, 205, 0}
\definecolor{cellblue}{RGB}{54, 125, 189}
\definecolor{codegreen}{rgb}{0,0.6,0}
\definecolor{codegray}{rgb}{0.5,0.5,0.5}
\definecolor{codepurple}{rgb}{0.58,0,0.82}
\definecolor{backcolour}{rgb}{1.0,1.0,1.0}
\lstdefinestyle{mystyle}{
    backgroundcolor=\color{backcolour},
    commentstyle=\color{codegreen},
    keywordstyle=\color{magenta},
    numberstyle=\tiny\color{codegray},
    stringstyle=\color{codepurple},
    basicstyle=\ttfamily\scriptsize,
    breakatwhitespace=false,
    breaklines=true,
    captionpos=b,
    keepspaces=true,
    numbers=left,
    numbersep=5pt,
    showspaces=false,
    showstringspaces=false,
    showtabs=false,
    tabsize=2
}
\newcommand{\tocite}[1]{{\color{red} [TO CITE]}}
\newcommand{\methodname}{AniDoc}
\newcommand{\method}{\texttt{\methodname}\xspace}
\title{\methodname: Animation Creation Made Easier}

\author{Yihao Meng$^{1,2}$ \and
Hao Ouyang$^{2}$ \and 
Hanlin Wang$^{3,2}$ \and
Qiuyu Wang$^{2}$ \and
Wen Wang$^{4,2}$ \and
Ka Leong Cheng$^{1,2}$ \and
Zhiheng Liu$^{5}$ \and
Yujun Shen$^{2}$ \and 
Huamin Qu$^{\dagger, 1}$ \and \hspace{0.9\linewidth}
\and $^{1}$ HKUST \and $^{2}$ Ant Group \and $^{3}$ NJU \and $^{4}$ ZJU \and $^{5}$ HKU
}

\begin{document}
\twocolumn[{
    \renewcommand\twocolumn[1][]{#1}
    \maketitle
    \begin{center}
      \vspace{-2pt}
      \includegraphics[width=\textwidth]{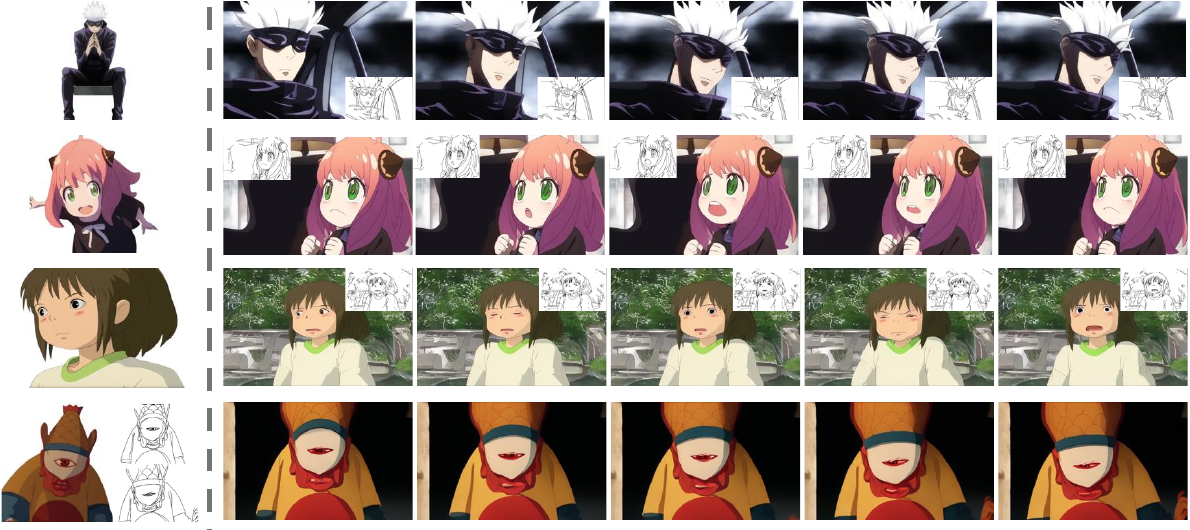}
      \vspace{-20pt}
      \captionsetup{type=figure}
      \caption{\method colorizes a sequence of sketches based on a character design reference with high fidelity, even when the sketches significantly differ in pose and scale.  Additionally, the model supports sparse sketch inputs, enabling effective interpolation and high-quality colorization simultaneously, as shown in the last row. }
      \label{fig:teaser}
      \vspace{10pt}
    \end{center}
}]

\maketitle

\begin{abstract}

The production of 2D animation follows an industry-standard workflow, encompassing four essential stages: character design, keyframe animation, in-betweening, and coloring.
Our research focuses on reducing the labor costs in the above process by harnessing the potential of increasingly powerful generative AI.
Using video diffusion models as the foundation, \method\footnote{``Doc'' is one of the seven dwarfs in \textit{Snow White and the Seven Dwarfs}, the first animated feature film produced by Disney.}
emerges as a video line art colorization tool, which automatically converts sketch sequences into colored animations following the reference character specification.
Our model exploits correspondence matching as an explicit guidance, yielding strong robustness to the variations (\textit{e.g.}, posture) between the reference character and each line art frame.
In addition, our model could even automate the in-betweening process, such that users can easily create a temporally consistent animation by simply providing a character image as well as the start and end sketches.
Our code is available at:
\textcolor{magenta}{\href{https://yihao-meng.github.io/AniDoc_demo}{https://yihao-meng.github.io/AniDoc\_demo}}.

\iffalse
Automating the colorization of line art in videos is crucial for streamlining animation production workflows and reducing labor costs. However, challenges such as misalignment between character design art and line art sketches, as well as the need for temporal consistency, hinder automation efforts. Previous methods often require manually colored keyframes and dense line art guidance, increasing the artist's workload and suffering from color information leakage due to non-binarized sketch conditioning. We propose a novel all-in-one model that leverages priors from video diffusion model to automate the colorization process. Our approach introduces an explicit correspondence mechanism with an injection module to align color information from reference images to input sketches, enhancing color accuracy. A two-stage training strategy learns to interpolate between keyframes, reducing the need for sketching intermediate frames. By conditioning on binarized sketches and employing data augmentation techniques, we improve training stability. Our method demonstrates superior quantitative and qualitative results, offering an effective solution for automatic line art video colorization and advancing the efficiency of animation production.
\fi

\end{abstract}

\section{Introduction}\label{sec:intro}

The animation industry, particularly in the realm of 2D anime production, relies heavily on the meticulous process of coloring line art to bring characters and scenes to life. Colorization of line art~\cite{yan2024colorizediffusion,huang2022unicolor,zou2024lightweight, Fourey_Tschumperlé_Revoy_2018, Furusawa_Hiroshiba_Ogaki_Odagiri_2017, Parakkat_Memari_Cani_2022, Sýkora_Dingliana_Collins_2009} in videos is a critical task that not only adds aesthetic value but also enhances the storytelling experience by conveying emotions and actions vividly. Automating this process holds significant potential to streamline production workflows, reduce labor costs, and accelerate content creation, meeting the growing demand for high-quality animated content.

\begin{figure}[t]
    \centering
    \includegraphics[width=\columnwidth]{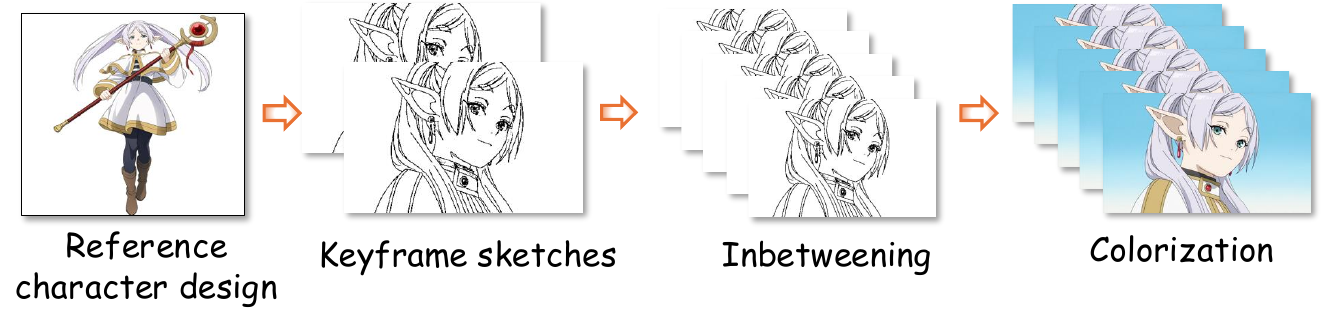}
    \caption{Illustration of the workflow of 2D animation production.}
    \label{fig:industry}
    \vspace{-10pt}
\end{figure}

In the current anime production pipeline, artists typically begin with character design sheets that define the visual attributes of characters. These designs are then translated into keyframe  sketches — crucial frames that outline the primary poses and movements in a scene. Next, artists create in-betweening sketches, which are the frames drawn between the keyframes to define the detailed movement and transition~\cite{zhu2024thin, siyao2021AnimeInterp}. Traditionally, these frames are manually colored, a time-consuming task that involves careful attention to ensure consistency with the original character designs.~\cref{fig:industry} illustrates each step of this pipeline. Our work aligns seamlessly with this pipeline, aiming to automate the colorization process while maintaining fidelity to the original character designs and ensuring temporal consistency across frames.

However, automating line art colorization~\cite{li2022eliminating,zhang2021user} presents several challenges. One primary difficulty lies in the mismatch between the character design and the line art sketches, where the angles, proportions, and poses in the design may not align with those in the keyframe sketches. Additionally, achieving temporal consistency is crucial; colorizing each frame individually can lead to flickering or inconsistencies, detracting from the viewer's experience~\cite{maejima2019graph, zhou2021cocosnet,carrillo2023diffusart}. Previous approaches~\cite{huang2024lvcd, tcvc,yu2024ACOF} have attempted to address these challenges but with limitations. They often assume the availability of colorized versions of keyframes and rely on dense line art guidance. This assumption significantly increases the workload on artists, as it requires manual coloring of multiple keyframes and detailed line art inputs, making the process tedious and labor-intensive. Moreover, some methods suffer from color information leakage due to their training pipelines. Specifically, they use non-binarized sketches extracted from color images using neural networks for training, unintentionally incorporating color information from the original images into the sketches. This information leakage undermines the practicality of these methods, as real-world sketches do not contain such implicit color information—a concern we analyze further in our methodology.

To overcome these challenges, we propose a novel all-in-one model that streamlines the colorization process within a single framework. Our model leverages the priors from pretrained diffusion-based video generation models ~\cite{svd,sd}, harnessing their ability to capture temporal dynamics and visual coherence. The key designs of our approach are as follows: \textbf{Correspondence-guided Colorization}: We address the misalignment between the reference character design and the input line art sketches by incorporating an explicit correspondence mechanism. The injection module is designed to integrate color and style information from the reference into the line art, effectively improving color accuracy and consistency. 
% This module aligns features between the reference and the input, ensuring that the colorization adheres closely to the original designs. 
\textbf{Binarization and Background Augmentation}: To reflect real usage scenarios, we binarize the condition sketches, forcing the model to truly learn to extract color information from the reference character design, rather than relying on recovering color information leaked from the non-binarized sketches. This constraint poses additional challenges for the model to accurately colorize the line art. To mitigate the instability during training due to this reduced information, we incorporate background augmentation strategy which greatly improve the colorization result. \textbf{Sparse Sketch Training}: Our model employs a two-stage training strategy that first learns the colorization ability and then removes the intermediate sketches to learn the interpolation ability.  By learning the interpolation between keyframes, our model maintains temporal consistency without extensive human intervention. 
% This approach reduces the need for manual drawing of intermediate sketches, significantly lowering the production cost and artist workload.

 Our method demonstrates superior performance both quantitatively and qualitatively compared to existing approaches. It effectively colorizes line art sketches in videos, maintaining high fidelity to the reference character designs and ensuring temporal consistency across frames. Moreover,  we demonstrate that a single reference character image can be used to colorize sketches from different segments featuring the same character, even when these sketches differ significantly in scale, pose, and action from the reference design. Our work represents a significant step toward automated, efficient, and artistically consistent animation production, with potential applications extending beyond anime to various forms of digital art and media.

% In summary, our contributions are threefold: We introduce a novel all-in-one model that automates the colorization of line art in videos, aligning with the existing anime production pipeline and reducing the need for manual intervention. We design an explicit correspondence mechanism with an injection module to accurately align and transfer color information from reference character design to input sketches, enhancing color accuracy.
% We identify an information leakage problem inherent in existing coloring methods—a critical yet previously unrecognized issue. By binarizing sketches, we ensure that our trained model can be effectively applied to artist hand-drawn sketches in real-world scenarios.

% start from anime production, tedious, 

% 然后讲other methods' problem: reference  一定要是首帧, binary, too dense

% 然后讲我们怎么解决. 

% evaluation

% summarize contribution

\section{Related Work}\label{sec:related}

\subsection{Line Art Image Colorization}

Line art colorization~\cite{zhang2023adding, cao2021line, kim2019tag2pix, dou2021dual, zhang2021user, zhang2018two, liu2018auto, sangkloy2017scribbler, zhang2021line} differs from natural image colorization, as it lacks an illuminance channel and only contains structural information, offering more flexibility. Traditional methods~\cite{Qu_2006,S_2009} rely on users manually adding color to specific regions. The advent of deep learning has advanced this field, with techniques like color hint points~\cite{Zhang_2018}, color scribbles~\cite{Ci_2018}, text tags~\cite{Kim_2019}, and natural language prompts~\cite{Zou_2019}. Reference-based colorization, where users provide a reference image to guide the coloring, has also gained popularity. Methods like~\cite{Sato_2014} use segmented graphs, while Chen et al.~\cite{chen2020active} employ active learning, and~\cite{Cao_yu_2022} apply attention networks. AnimeDiffusion~\cite{cao2024animediffusion} introduces the first diffusion-based reference-based framework for anime face colorization. However, these methods colorize sketch images separately, struggling with temporal coherence when colorize a sequence of sketches.

\subsection{Reference-based Line Art Video Colorization}

Several approaches extend reference-based colorization to videos. Given an input of reference image, these methods colorize the corresponding sketch sequence based on the reference image's color information. LCMFTN~\cite{Zhang_2020} trains a model using animation frame pairs but lacked temporal coherence. TCVC~\cite{tcvc} uses previously colorized frames to maintain short-term consistency, but errors accumulate over time. TRE-Net~\cite{wang2023TRE-Net} mitigates this by using both the first frame and previously generated frames. ACOF~\cite{yu2024ACOF} propagates colors based on optical flow but requires refinement. The most recent work LVCD~\cite{huang2024lvcd} proposes the first video diffusion model to colorize sketch video, but suffer from the issue of non-binarized sketch information leakage, limiting its real-world applicability. Most importantly, all of these methods require the color version of the first frame of each video clip, while in the actual anime production workflow, colorists typically only receive the character design image and need to use that image to colorize different clips featuring the character.

\subsection{Video Interpolation}

Unlike reference-based colorization, video interpolation~\cite{wang2024framer, hvfi, danier2024ldmvfi, jain2024video, superslomo2018, huang2020rife} aims to generate in-between frames from both the first and last frames. SparseCtrl~\cite{Guo_2023} and SEINE~\cite{chen2023seine} extend video interpolation using pretrained text-to-video diffusion models and additional image conditions. Shi et al.~\cite{shi2020deep} adapt video interpolation for cartoon animation with temporal constraint networks.  AnimeInterp~\cite{siyao2021AnimeInterp} interpolates middle frames by warping with predicted flows. To improve animation quality, EISAI~\cite{chen2022eisai} uses forward warping to prevent artifacts. ToonCrafter~\cite{xing2024tooncrafter} introduces a diffusion-based video interpolation model.
Zhu et al.~\cite{zhu2024thin} proposes a thin-plate spline-based interpolation module for animation sketch inbetweening. We aim at an all-in-one model that simultaneously performs automatic interpolation and colorization for anime. However, more precise interpolation methods~\cite{wang2024framer}, can be integrated to achieve enhanced control over sketch manipulation.

\begin{figure*}[t] 
    \centering
    \includegraphics[width=\linewidth]{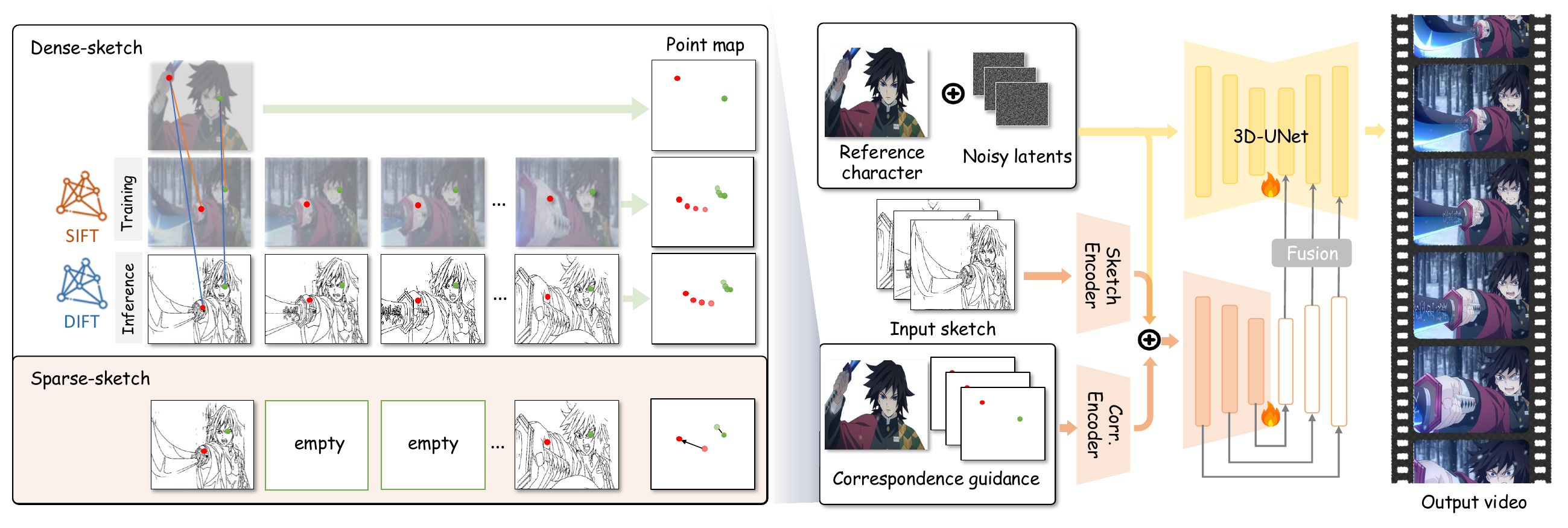} %
\caption{Overview of \method pipeline. We adopt a two-stage training strategy. In the \textit{dense-sketch} training stage, we explicitly extract matching keypoints pairs between the reference image and each frame of the training video, constructing point maps to represent the correspondences. In the \textit{sparse-sketch} training stage, we remove the intermediate frame sketches and use the matching points from the start and end frames to interpolate point trajectories, which guide the generation of the intermediate frames.}
    % \vspace{-2ex} 
\end{figure*}

\section{Method}\label{sec:method}
We formally define the problem of line art video colorization with reference images. Given
a reference image \( I_{\text{ref}} \) that encapsulates the desired color and style of the character and a sequence of binarized line art sketches \( \{ S_t \}_{t=1}^T \), where \( S_t \) is the sketch at time frame \( t \), our objective is to generate a sequence of colorized frames \( \{ I_t \}_{t=1}^T \) such that:
\begin{enumerate}
    \item Each frame \( I_t \) is a colorized version of the sketch \( S_t \).
    \item The colorization is consistent with the character design.
    \item The sequence \( \{ I_t \} \) is temporally coherent, ensuring smooth transitions without flickering artifacts.
\end{enumerate}
Formally, we aim to learn a function \( f \) that maps the sketches and the reference image to the colorized frames:
\begin{equation}
    \{I_t\}_{t=1}^T = f(\{S_t\}_{t=1}^T, I_{\text{ref}}).
\end{equation}

% , \quad \text{for } t = 1, 2, \dots, T

% We then analyze the existing issues for the animation creation in real applications.
% such as LVCD and ToonCrafter 

\subsection{Motivation and Pipeline Design} 
% Previous state-of-the-art approaches exhibit considerable limitations when handling real-world animation production scenarios. To address these issues, we analyzed the shortcomings of these methods and designed modules to overcome them. 

We summarize the shortcomings of state-of-the-art approaches when handling real-world animation production scenarios and design modules to overcome them.

\noindent \textbf{Mismatch Between Character Design and Input Sketches.}  
Existing methods rely on the assumption that the reference image is strongly pixel-aligned with the first frame of the sketch sequence. When the reference provided is not the first-frame image, but instead a character design from a different angle, the model struggles to correctly match the colors and details, as shown in~\cref{fig:comparison}. This limitation necessitates manual coloring for each clip's keyframes in the animation, which is labor-intensive and counteracts the benefits of automation.  To address this, we aligns and transfers color information from reference character design to input sketches with correspondence-guided control module, as described in~\cref{subsec:corr}.

% As shown in Figure 2(a), when using previous methods on input sketches that do not align closely with the character design, the colorization output is erroneous. The colors are misplaced, and the visual fidelity to the character design is lost. 

\begin{figure}[ht]
    \centering
    \includegraphics[width=\columnwidth]{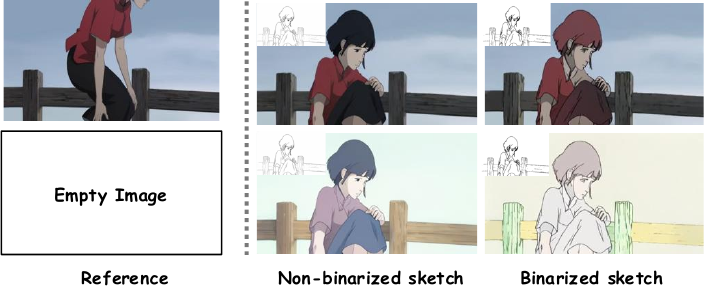}
    \caption{Illustration of color leakage issue in non-binarized sketch. For previous video colorization method~\cite{huang2024lvcd}, when given non-binarized sketch, even if the reference is an empty image, it can still generate colorized results with similar color pattern to the ground truth. After binarizing the sketch, the colorization results degrade significantly. }
    \label{fig:leakage}
    \vspace{-10pt}
\end{figure}
\noindent \textbf{Degradation with Binarized Sketches.}   Previous methods often rely on sketches extracted from color images using learned neural networks. These non-binarized sketches contain unintended color information leaks from the original color images,  although invisible to the human eye. When training with these non-binarized sketches, the model tends to learn how to recover these hidden color information rather than correctly learning to find the corresponding parts in the reference to color the sketch, resulting in models that perform poorly when applied to binarized sketches. As depicted in~\cref{fig:leakage}, when previous methods are applied to binarized sketches, the outputs suffer from severe degradation. The colorization is inaccurate, and the visual quality is significantly reduced compared to when non-binarized sketches are used. To address this issue, we mirror the real production conditions by adopting the binary sketch in training and apply background augmentation to enhance the robustness as in \cref{subsec:bin}.

\noindent \textbf{Reliance on Dense Sketches as Conditions.}  Previous methods often require dense sketches to maintain temporal consistency. Manually drawing  inbetweening sketches in an animation is costly.  We aim to use only \( S_{1}, S_{T} \) to further improve the scalability of automatic creation and propose the sparse sketch training scheme in \cref{subsec:sparse}.

% The automation of lineart colorization in animation faces significant challenges that existing methods struggle to overcome.

 % In traditional animation workflows, character designs provide detailed color and style references for animators. However, the lineart sketches used in keyframes often differ in pose, scale, and details from the character design sheets.

% In real animation production, the sketches available are typically binarized line drawings without grayscale shading or additional details. 

\noindent \textbf{{Pipeline Design}.} Following Stable Video Diffusion (SVD)~\cite{svd}, our main architecture consists of a denoising 3D U-Net designed for video generation. The reference image latent is duplicated across the number of frames and concatenated along the channel dimension with the noisy latent so that the reference image information can be integrated into the colorization process through multiple self-attention layers in the 3D U-Net encoder. To inject the correspondence between the reference character design and the sketch, we explicitly extract corresponded keypoints and construct point maps. The correspondence information and sketch information are then encoded by a 3D U-Net dual branch similar to~\cite{zhang2023controlnet} , and injected into the main branch as control signals.
To construct training pairs of reference character designs and corresponding sketch sequences, we select long videos from the Sakuga-42M dataset~\cite{pan2024sakuga}, where the characters naturally undergo rich transformations, including changes in position, posture, angle, scale, etc., which align with the needs of our model. 

% including temporal attention and 3D convolutional layers, along with their associated weights. 

% This duplicated encoder is used to encode the correspondence information and sketch information as guidance signals, which is then injected into the decoder of the main branch at various layers using zero convolution. 

% Considering that the prior of SVD itself does not completely align with our requirements, as shown in~\cref{fig:svd_prior}, we do not just fine-tune the parameters of ControlNet during training, but instead fine-tune all parameters including the main branch together. 

 % choosing reference images from the beginning of the video as the character design images. For the training video, we extract frames from the latter half of the video and extract sketches. In these long videos, 

The training objective for our model is as follows:

\begin{equation}
\mathcal{L}=\mathbb{E}_{z_t, z^0, t, \epsilon}\left[\left\|\epsilon-\epsilon^{c}_{\theta}\left(z_t; t, z^0, c_{sketch},c_{corr}\right)\right\|^2\right],
\label{eq:conditonal_denoising}
\end{equation}
where \( z^0 \) denotes VAE-encoded latent feature of the reference image, \( c_{sketch}\) means the sketch control signal, \(  c_{corr}\) means the correspondence control signal, and \( \epsilon_\theta^c \) is the combination of the denoising U-Net and the control branch.

% \label{subsec:corr}
\subsection{Correspondence-guided Colorization}
\label{subsec:corr}
% \begin{figure}[ht]
%     \centering
%     \includegraphics[width=\columnwidth]{pdfs/svd_prior.pdf}
%     \caption{In the early training stage (1w step), the video generation model produces static videos that closely resemble the given reference design. }
%     \label{fig:svd_prior}
% \end{figure}

% As an image-to-video model, SVD (Stable Video Diffusion)~\cite{svd} inherently possesses the ability to extract information from an input image to generate a video. However, during training, we observed that the strong prior in SVD restrict the first frame to be the same with the input reference image, as shown in~\cref{fig:svd_prior}.

%  In our formulation, the input image is not the first frame of the video, but rather a reference character design from a different viewpoint. The model needs to query colors from this reference image, while the structure information should align with the given sketch list. This conflicting prior makes training the model significantly more challenging.

% To better establish the relationship between the reference character design and the sketch, reduce the learning difficulty for the model, and improve the fine-grained details, we propose Correspondence Matching Module. This module explicitly injects the matching relationships between the reference image and the sketch, enabling the model to better query and color the correct areas.

% \subsection{Keypoint Matching and Point Map Construction}

% \noindent \textbf{{Keypoint Matching and Point Map Construction}.}
During training, we use an off-the-shelf keypoint matching method LightGlue~\cite{lightglue} with SIFT descriptor~\cite{lowe2004sift} to extract matching keypoints between the reference image and the first frame of the training video. 
The matched keypoints are denoted as $\left\{(x_{\text{ref}}^i,y_{\text{ref}}^i),(x_1^i,y_1^i)\right\}_{i=1}^n $, where \(n\) refers to the number of matching pairs.
Based on these matched keypoints, we construct a point map pair \( P_1=(P_{\text{ref}}, P_{1,\text{sketch}}) \), representing the matching correspondence of the reference image and the first frame. 
Each point map is in the size of \( H \times W \) (same as the image resolution), where the coordinates corresponding to the matched keypoints are marked with the same integer label:
\begin{equation}
P_{\text{ref}}(x_{\text{ref}}^i,y_{\text{ref}}^i) = P_{1,\text{sketch}}(x_1^i,y_1^i) = i.
\end{equation}
For unmatched pixels, the value is set to 0.

 % During training, we randomly select keypoints number between 0 and 50, which helps make the model more robust to variations in the number of keypoints.

% \[
% P_{\text{ref}}(x, y),P_1(x,y) = 
% \begin{cases}
% i    & \text{if there is the \(i^{\text{th}}\) matching keypoint at } (x, y) \\
% 0    & \text{otherwise}
% \end{cases}
% \]

% \[
% \begin{aligned}
% P_{\text{ref}}(x, y), P_1(x,y) = 
% \begin{cases}
% i    & \text{if there is the \(i^{\text{th}}\) matching keypoint at } (x, y) \\
% 0    & \text{otherwise}
% \end{cases}
% \end{aligned}
% \]

% \begin{multline}
% P_{\text{ref}}(x, y),P_{1_{\text{sketch}}}(x,y) = \\
% \begin{cases}
% i    & \text{if there is the \(i^{\text{th}}\) matching keypoint at } (x, y) \\
% 0    & \text{otherwise}
% \end{cases}
% \end{multline}

% \noindent \textbf{{Gaussian Enhancement for Keypoints}.}

We then employ Co-Tracker~\cite{karaev2023cotracker} to track the movement of keypoints $\left\{(x_1^i,y_1^i)\right\}_{i=1}^n $ in each video frame and construct correspondence point maps in the same way. As a result, we obtain a point map \( P_{\text{seq}} \) = \(\{P_t\}_{t=1}^T\) $\in \mathbb{R}^{2 \times T \times H \times W}$, which explicitly encodes the correspondence between the reference image and each sketch in the training video.

% Since LightGlue typically extracts keypoints at the edges of the image, we do not want the model to consider only the color information at these keypoints when querying the reference image. Instead, we want the model to focus on a small local area around each keypoint to capture the surrounding color information. we enhance the keypoints with a 2D Gaussian representation. 

% In this representation, pixels closer to the keypoint have higher importance, and thus larger weights, while pixels farther away still have smaller but non-zero weights. 
 % Using the  keypoints coordinates in each frame, we construct Gaussian map \( G_{seq} \). This map is then concatenated with the corresponding point map \( P_{seq} \), forming a composite point map. While the point map and Gaussian map encode only spatial relationships, we further concatenate the reference character design with them as an additional input signal to provide the color information.
 
 We concatenate the point map \(P_{\text{seq}}\) with the reference image \(I_{\text{ref}}\) together as explicit correspondence guidance information and integrate it into the control branch. Specifically, for a given keypoint \( (x_t^i,y_t^i) \) in sketch frame \( S_t \), the model can use the point map \(P_t\) to obtain the corresponding location \( (x_{\text{ref}}^i,y_{\text{ref}}^i) \) in the reference image \( I_{\text{ref}} \) and extract corresponding color information at that location. Thus, our video generation process can be represented as:
 \begin{equation}
     \{I_t\}_{t=1}^T = D (I_{\text{ref}}, E(\{S_t\}_{t=1}^T, \{P_{t}\}_{t=1}^T, I_{\text{ref}})  ),
 \end{equation}
 where \(D\) refers to the denoising process of the 3D diffusion U-net, \(E\) refers to the control branch encoder.
% \[
% \text{Composite Point Map}_{t} = P_{t} \oplus G_{t},
% \]
%  where \( \oplus \) denotes concatenation.

%  After constructing point map and Gaussian map, we further concatenate the reference character design with them  as input to ControlNet.
% The motivation behind this design is that ControlNet provides precise pixel-to-pixel control over the generated video, while the point map and Gaussian map encode only spatial relationships and not color information. By concatenating the reference image, which contains pixel-to-pixel color information, the model can use the spatial correspondences to query the correct regions in the reference image for color. Specifically, for a given keypoint \( k_{i,t} \) in frame \( t \), the model can use the corresponding location in the reference image, obtained through the point map, and query the color at that location using the concatenated reference image. This allows the model to achieve pixel-to-pixel color control over the video generation process.

% \begin{figure}[ht]
%     \centering
%     \includegraphics[width=\columnwidth]{pdfs/dift_matching.pdf}
%     \caption{Semantic feature can effectively find matching keypoints between reference color image and binarized sketch.  }
%     \label{fig:dift_matching}
% \end{figure}

\noindent \textbf{{Semantic Keypoint Matching During Inference}.} During training, we apply LightGlue~\cite{lightglue} with SIFT descriptor~\cite{lowe2004sift} for keypoint selection and matching between the reference image and the training video frames due to its fast speed. However, during inference, we do not have the ground truth color image to extract corresponding keypoints. Methods like the SIFT descriptor, which are focusing on low-level image feature, fail to correctly match keypoints between the color reference image and the sketch due to the large domain gap. One recent work DIFT~\cite{dift} have found that features extracted by diffusion models can achieve semantic-level matching. Thus, during inference, we first extract keypoints in the reference character design using X-Pose~\cite{x-pose}, and find the matching keypoints in the given sketch using semantic feature DIFT.

\subsection{Binarization and Background Augmentation}
\label{subsec:bin}

% In real animation production, sketches provided to colorists are typically binarized line drawings devoid of grayscale shading or hidden color information. To simulate these real-world conditions, we apply a binarization process to the extracted sketches during training. Specifically, after extracting the raw sketch \( S_{\text{raw}} \) from the original image, we set the pixel value greater than 200 to 255 and otherwise 0. This process converts the sketch into a binary image \( S_{\text{bin}} \), where the lines are represented by black pixels (0) and the background and white regions are represented by white pixels (255).

In real animation production, sketches provided to colorists are typically binarized line drawings without grayscale shading or hidden color information. To simulate this real-world condition, we apply a binarization process to the extracted sketches during training by setting pixel values greater than 200 to 255 and others to 0. This converts the sketch \( S_{\text{raw}} \) into a binary image \( S_{\text{bin}} \), with black lines (0) and white background (255).

% \[
% S_{\text{bin}}(i,j) = 
% \begin{cases}
% 255, & \text{if } S_{\text{raw}}(i,j) > 200, \\
% 0, & \text{otherwise},
% \end{cases}
% \]
% where \( (i,j) \) denotes the pixel coordinates.

Using binarized sketches as conditioning inputs poses significant challenges for the model. One primary issue is the ambiguity between the background and large white regions in the foreground. Both are represented by the same pixel value (255), making it difficult for the model to distinguish. This ambiguity can lead to confusion during colorization, resulting in erroneous outputs where the model may incorrectly color background regions or fail to accurately color foreground elements. Such failure cases are evident in our ablation studies (refer to~\cref{sec:ablation}).

To address this problem, we enhance the training process through background augmentation. Specifically,  we randomly remove the background of the reference image with a 50\% probability during training process, using an off-the-shelf background removal model~\cite{u2net}. This forces the model to learn to distinguish between the foreground and background regions. In the foreground region, the model learns to extract color information from the correct areas of the reference character design. In the background region, the model is required to rely more heavily on its internal generative prior, enabling it to produce a background that is coherent with the foreground character.

\subsection{Sparse Sketch Training}
\label{subsec:sparse}

To further reduce the necessity of drawing intermediate sketches for animations with simple motions, we introduce a sparse sketch training strategy in a two stage way. In the first stage, the training is conducted with all frame sketches available. We hope that the model learns how to correctly extract information from the point map in this stage, which will guide the training in the next stage.

After completing the first stage training, we remove the sketch condition for intermediate frames and use keypoints information to guide the interpolation.
Inspired by \cite{wu2025draganything}, we additionally transform the keypoint coordinates into a Gaussian heatmap \(G_{t}\)  which is more suitable for trajectory control, and concatenate \(G_{t}\) with the original point map \(P_{t}\) together.
Specifically, the model is conditioned on the start and end sketches, \( S_1\) and \(  S_T \), as well as the pointmap \(P_{t} \oplus G_{t}\) at each frame: 
\begin{equation}
    \{I_t\}_{t=1}^T= D (I_{\text{ref}}, E(S_1,S_T , \{P_{t} \oplus G_{t}\}_{t=1}^T, I_{\text{ref}})).
\end{equation}

Noted that during training the point trajectories are obtained through point tracking in the training video using Co-Tracker~\cite{karaev2023cotracker}, while during inference we extract matching keypoints pairs between the start and end sketches and linearly interpolate the intermediate point trajectories.

In this sparse-sketch training stage, we randomly select up to 5 keypoints, corresponding to 5 trajectories. The selection probability is determined by the magnitude of the motion, with trajectories having larger motion being more likely to be selected.
Guided by the keypoint trajectories, our model can produce smooth intermediate colorized frames with only sparse sketch inputs.
% \[
% I_t = D\left( M\left( \mathbf{0}, E_{\text{ref}}(I_{\text{ref}}) \right), P_t \right), \quad \text{for } t = 2,\dots,T-1.
% \]

% During inference, we generate intermediate keypoints trajectories by interpolating the corresponding points of the start and end keyframes, enabling the model to produce colorized frames with only sparse sketch inputs.

 % This approach significantly lowers the manual drawing effort while maintaining high-quality and temporally coherent colorization results.

% To address this issue, we introduce the Correspondence-Guided Control Module. This module establishes explicit correspondences between the reference character design and the input sketches, even when there are significant differences in pose or details. By aligning the features of the reference and input, our model accurately transfers color information, ensuring that the colorization adheres to the intended design regardless of the initial misalignment.
% Motivation:

\section{Experiments}\label{sec:exp}

\begin{figure*}[t] 
    \centering
    \includegraphics[width=\textwidth]{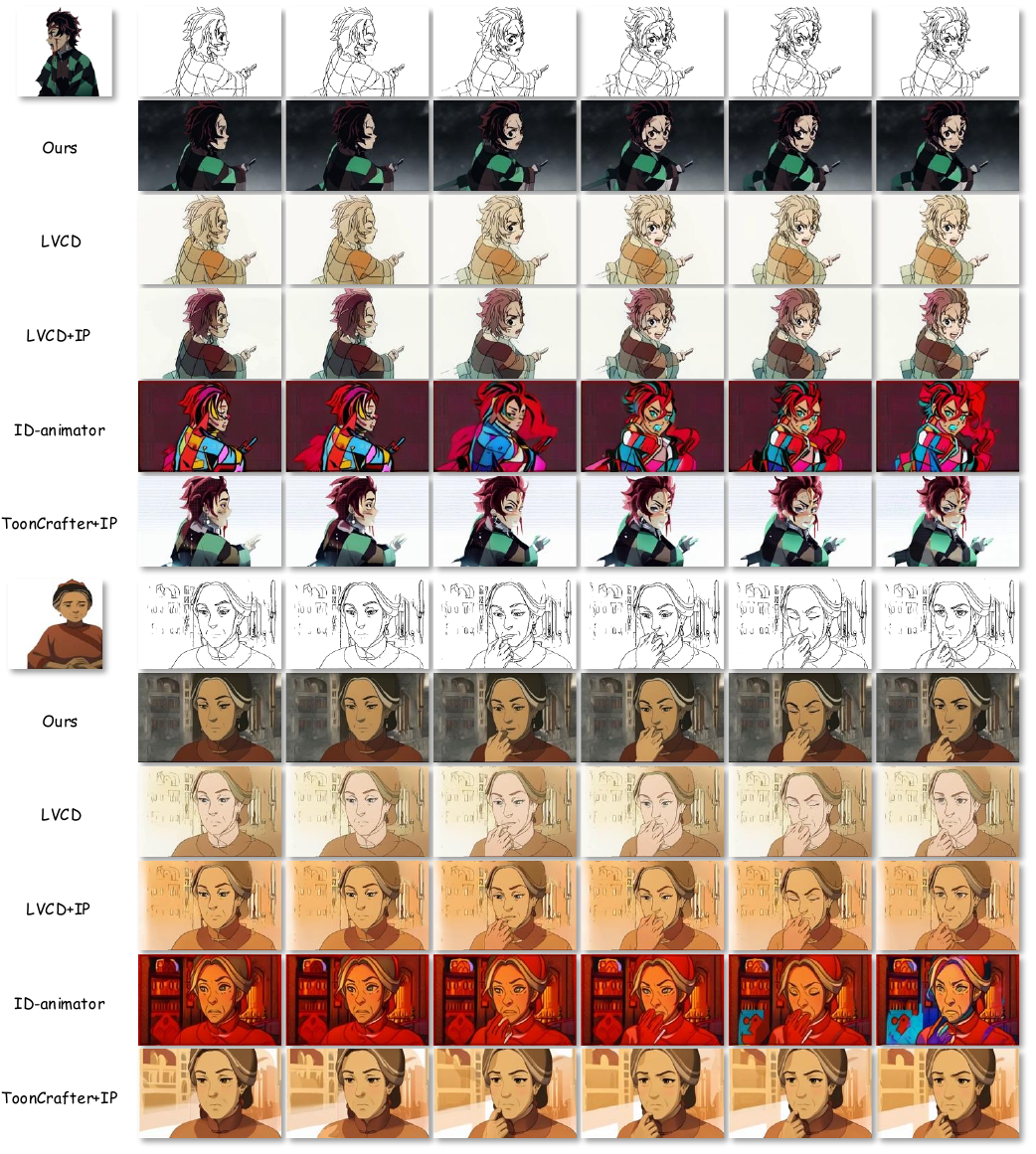} 
    \caption{Visual comparison of reference-based colorization  with four methods LVCD~\cite{huang2024lvcd}, LVCD+IP-Adapter~\cite{ye2023ip}, ID-animator~\cite{he2024id}, ToonCrafter~\cite{xing2024tooncrafter}.} 
    \vspace{-2ex} 
    \label{fig:comparison} 
\end{figure*}

\begin{figure}[ht]
    \centering
    \includegraphics[width=\columnwidth]{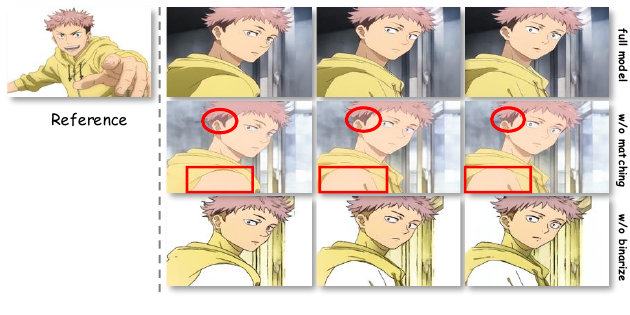}
    \caption{Ablations on each component. ``\textit{w/o} matching'' indicates without the corresponding matching module, ``\textit{w/o} binarize'' indicates without binarization and background augmentation.}
    \label{fig:ablation}
    \vspace{-10pt}
\end{figure}
\subsection{Implementation Details}

Our method is built upon SVD and is trained on the Sakuga-42M~\cite{pan2024sakuga} dataset, which comprises a large number of anime clips with diverse styles. 
To create reference images and sketch videos with large differences, we exclude clips with fewer than 50 frames, ultimately retaining around 150k video clips. We set the interval between the reference image and the first frame of the target video to 32 frames, with the target video length being 14 frames. During the first training stage, where the generation is conditioned on per-frame sketches, we simultaneously fine-tune all parameters of both the U-Net and ControlNet, including both the spatial and temporal attention layers. This is done for 100k steps using the AdamW optimizer with a learning rate of \( 1 \times 10^{-5} \), at a resolution of $256\times256$  due to GPU memory constraints. Subsequently, we freeze other layers and fine-tune the spatial layers for an additional 10k steps at a resolution of $512\times320$. We randomly select up to 50 keypoints to enhance the model's robustness to varying numbers of keypoints. During the sparse sketch training stage, we remove the middle frames' sketches and further fine-tuned all parameters for 100k steps.  The training is conducted on 16 NVIDIA A100 GPUs with a total batch size of 16, and the entire training process takes 5 days.

\subsection{Comparison}

To comprehensively evaluate the coloring ability of our model, we randomly select 200 clips from 10 different eras and styles of anime to construct the test set. We select corresponding character design images (without background) as reference images.

We compare our proposed method with two state-of-the-art reference-based line art video colorization frameworks: LVCD~\cite{huang2024lvcd} and ToonCrafter~\cite{xing2024tooncrafter}, both of which are based on video diffusion models. Since LVCD's original setting requires the colorized version of the first sketch as a reference, we also compare it with the IP-Adapter~\cite{ye2023ip} + LVCD version. In this case, we first use a diffusion-based image colorization method IP-Adapter to colorize the first frame's sketch, and then use this colorized image as the reference for LVCD. For ToonCrafter, which requires color versions of both the start and end frames, we first use IP-Adapter to colorize the sketches of both frames. Additionally, we select a recent video personalization method, ID-animator~\cite{he2024id}, which excels at identity preservation in general domains and can achieve a similar function to colorization when combined with ControlNet~\cite{zhang2023controlnet}.
% We also compare our method with
% we select baselines from two distinct categories. The first category includes state-of-the-art video models for animation: \textbf{LVCD} and \textbf{ToonCrafter}. Since LVCD and ToonCrafter require a colorized version of the first frame, we use \textbf{IP-Adapter} to perform the initial colorization. The second category includes the recent reference-based video model for general domains: \textbf{ID-Animator}. We conduct both quantitative and qualitative analyses on the test set of Sakuga-42M.

\noindent\textbf{Qualitative Comparison.}
As shown in~\cref{fig:comparison}, our method produces significantly clearer textures and better preserves the character's identity. It performs especially well in scenarios with substantial differences between the reference character design and input sketches, where LVCD and ID-Animator fail to accurately colorize the sketches. Even when providing the colorized first frame using IP-Adapter for these baselines, our method still outperforms them in both visual quality and identity preservation.

\begin{table}
    \centering
    \caption{
        Quantitative comparison with existing baselines on reconstruction and generative metrics.
    }
    \label{tab:comparison}
    \vspace{-2pt}
    \SetTblrInner{rowsep=1.2pt}      % Row space.
    \SetTblrInner{colsep=2.0pt}      % Col space.
    \begin{tblr}{
        cells={halign=c,valign=m},   % Text alignment for all cells.
        column{1}={halign=l},        % Text alignment for the first column.
        hline{1,2,6}={1-6}{},        % Horizontal lines.
        hline{1,7}={1.0pt},          % Horizontal line width.
        % vline{2}={1-4}{},          % Vertical lines.
    }
\ Method &PSNR$\uparrow$& SSIM$\uparrow$&LPIPS$\downarrow$ &FID$\downarrow$&FVD$\downarrow$\\ 
    ID-Animator             & 15.61 & 0.3129 & 0.5151 & 158.16 & 677.61 \\
    LVCD                    & 15.77 & 0.6446 & 0.2580 & 121.98 & 584.33 \\
    LVCD + IP               & 16.52 & 0.6404 & 0.2639 & 118.28 & 496.45 \\
    ToonCrafter + IP         & 14.97 & 0.3983 & 0.4532 & 110.48 & 492.10 \\
    \method (Ours) & \textbf{19.23} & \textbf{0.7720} & \textbf{0.1704} & \textbf{54.33} & \textbf{230.18} \\
    \end{tblr}
    \vspace{-5pt}
\end{table}

\noindent\textbf{Quantitative Comparison.}
We evaluate the quality of the colorized animations in two  aspects: 1). Frame and Video Quality: we adopt Frechet Inception Distance~\cite{heusel2017fid} (FID) to measure image quality and assess video quality using Frechet Video Distance~\cite{unterthiner2018fvd} (FVD).
2). Colorization Accuracy: we measure the similarity of the colorized frames and the original animation frames using reconstruction metrics including PSNR, SSIM, and LPIPS. For all the metrics, we resize the frames to \(256 \times 256 \) and remove the background for every frame because the reference character design does not include background information.

As shown in~\cref{tab:comparison}, our method achieves the best scores across all metrics, indicating high-quality colorization results. For a more comprehensive assessment, we recommend reviewing the supplementary comparison videos.

\begin{figure*}[t] 
    \centering
    \includegraphics[width=\textwidth]{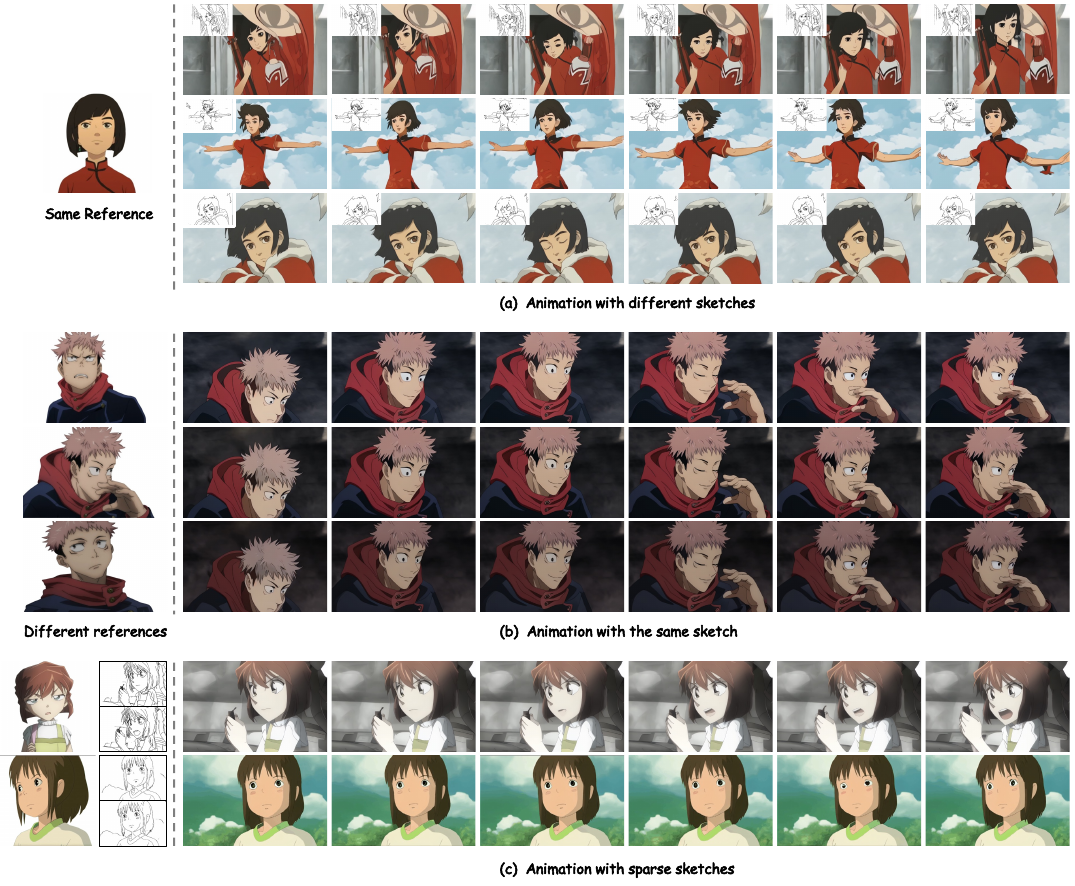} %
    \vspace{-2ex} 
    \caption{Illustration of the flexible usage of our model. Figure (a) shows the ability of using same reference to colorize different sketches. Figure (b) demonstrates the robustness to different references. Figure (c) shows the sparse-sketch generation results.}
    \label{fig:robustness}
    \vspace{-10pt}
\end{figure*}

\subsection{Flexible Usage}

We assess the flexibility of our model in three distinct settings, as depicted in~\cref{fig:robustness}.

\noindent\textbf{Same Reference with Varying Sketches.} By using the same reference, our model is able to generate consistent colorizations across different video clips, even when the sketches differ significantly in terms of pose or scale.

\noindent\textbf{Same Sketch with Different References.} When applying different reference images to the same sketch sequence, our method preserves the identity of the character while adapting the finer details, such as lighting and background, according to the distinct styles of the references.

\noindent\textbf{Sparse Input Sketches.} Thanks to our two-stage training strategy, 
 \method supports animation with sparse sketches. By using only the start and end sketches, the model effectively produces smooth and coherent animations.

\subsection{Ablation Study} \label{sec:ablation}

% \begin{table}[h]
% \small
% \renewcommand\tabcolsep{5.0pt}
% \centering
% \resizebox{\linewidth}{!}{
% \begin{tabular}[t]{rccccc}
% % \toprule
% \cmidrule{2-6}
% & PSNR$\uparrow$ & SSIM$\uparrow$ & LPIPS$\downarrow$ & FID$\downarrow$ & FVD$\downarrow$ \\ 
% \hline
% \textit{w/o} binary     & 15.41 & 0.6639 & 0.2471 & 94.89 & 503.23 \\
% \textit{w/o} matching   & 17.80 & 0.7068 & 0.2252 & 75.91 & 273.86 \\
% \hline
% \methodname (Ours) & \textbf{19.23} & \textbf{0.7720} & \textbf{0.1704} & \textbf{54.33} & \textbf{230.18} \\
% \hline
% % \bottomrule
% \end{tabular}
% }
% \caption{
%     Ablations on correspondence matching module and binarization+background augmentation.
% }
% \label{tab:ablations}
% % \vspace{-2ex}
% \end{table}

\begin{table}
    \centering
    \caption{
        Ablations on correspondence matching module and \textit{binarization + background} augmentation.
    }
    \label{tab:ablation}
    \vspace{-2pt}
    \SetTblrInner{rowsep=1.2pt}      % Row space.
    \SetTblrInner{colsep=2.0pt}      % Col space.
    \begin{tblr}{
        cells={halign=c,valign=m},   % Text alignment for all cells.
        column{1}={halign=l},        % Text alignment for the first column.
        hline{1,2,4}={1-6}{},        % Horizontal lines.
        hline{1,5}={1.0pt},          % Horizontal line width.
        % vline{2}={1-4}{},          % Vertical lines.
    }
\ Settings&PSNR$\uparrow$&SSIM$\uparrow$&LPIPS$\downarrow$&FID$\downarrow$&FVD$\downarrow$ \\ 
\textit{w/o} binary     & 15.41 & 0.6639 & 0.2471 & 94.89 & 503.23 \\
\textit{w/o} matching   & 17.80 & 0.7068 & 0.2252 & 75.91 & 273.86 \\
\method (Ours) & \textbf{19.23} & \textbf{0.7720} & \textbf{0.1704} & \textbf{54.33} & \textbf{230.18} \\
    \end{tblr}
    \vspace{-5pt}
\end{table}

We perform ablation studies~\cref{tab:ablation} on two key components.

\noindent\textbf{Effect of Correspondence Matching.}
Without the correspondence matching module, the model struggles to localize and transfer detailed color information accurately. As shown in the~\cref{fig:ablation}, the character's black sideburns have not been colored correctly, and the pink color of the hair has mistakenly been applied to the clothing. This miscoloring leads to poor preservation of the character's identity. Incorporating correspondence matching ensures precise alignment between the reference and the input sketches, significantly improving color accuracy.

\noindent\textbf{Effect of Background Augmentation.}
Without background augmentation during training, the model struggles to distinguish between the foreground (character) and the background. Consequently, it may generate frames where certain regions are incorrectly colorized as pure white or contain artifacts, due to the limited information provided by the binarized sketches. Incorporating background augmentation helps the model to better understand scene context, leading to more accurate and visually pleasing results.

\section{Conclusion}\label{sec:conclusion}
In this paper, we introduced a novel all-in-one model for automatic lineart video colorization that integrates seamlessly with existing anime production pipelines. Our approach tackles key challenges such as misalignment between character designs, limited information in binarized sketches and situations with only sparse sketches.  Comprehensive experiments demonstrate that our method outperforms state-of-the-art baselines in both quality and temporal consistency. For future work, we aim to incorporate interactive point control to handle subtle color variations and develop stronger, more efficient video models to support longer and higher-quality animation creation, further enhancing the efficiency and creativity of animation production.

% By continuing to refine our approach and address the remaining challenges, we aim to provide powerful tools that empower artists and studios to produce high-quality animated content with greater ease and creativity.

\newpage
{
\small
\bibliographystyle{ieeenat_fullname}
\bibliography{ref.bib}

\begin{thebibliography}{67}
\providecommand{\natexlab}[1]{#1}
\providecommand{\url}[1]{\texttt{#1}}
\expandafter\ifx\csname urlstyle\endcsname\relax
  \providecommand{\doi}[1]{doi: #1}\else
  \providecommand{\doi}{doi: \begingroup \urlstyle{rm}\Url}\fi

\bibitem[Blattmann et~al.(2023)Blattmann, Dockhorn, Kulal, Mendelevitch, Kilian, Lorenz, Levi, English, Voleti, Letts, Jampani, and Rombach]{svd}
Andreas Blattmann, Tim Dockhorn, Sumith Kulal, Daniel Mendelevitch, Maciej Kilian, Dominik Lorenz, Yam Levi, Zion English, Vikram Voleti, Adam Letts, Varun Jampani, and Robin Rombach.
\newblock Stable video diffusion: Scaling latent video diffusion models to large datasets.
\newblock \emph{CoRR}, abs/2311.15127, 2023.

\bibitem[Cao et~al.(2021)Cao, Mo, and Gao]{cao2021line}
Ruizhi Cao, Haoran Mo, and Chengying Gao.
\newblock Line art colorization based on explicit region segmentation.
\newblock In \emph{Computer Graphics Forum}, pages 1--10, 2021.

\bibitem[Cao et~al.(2023)Cao, Tian, and Mok]{Cao_yu_2022}
Yu Cao, Hao Tian, and P.~Y. Mok.
\newblock Attention-aware anime line drawing colorization.
\newblock In \emph{{IEEE} International Conference on Multimedia and Expo, {ICME} 2023, Brisbane, Australia, July 10-14, 2023}, pages 1637--1642. {IEEE}, 2023.

\bibitem[Cao et~al.(2024)Cao, Meng, Mok, Lee, Liu, and Li]{cao2024animediffusion}
Yu Cao, Xiangqiao Meng, PY Mok, Tong-Yee Lee, Xueting Liu, and Ping Li.
\newblock Animediffusion: anime diffusion colorization.
\newblock \emph{IEEE Transactions on Visualization and Computer Graphics}, 2024.

\bibitem[Carrillo et~al.(2023)Carrillo, Cl{\'e}ment, Bugeau, and Simo-Serra]{carrillo2023diffusart}
Hernan Carrillo, Micha{\"e}l Cl{\'e}ment, Aur{\'e}lie Bugeau, and Edgar Simo-Serra.
\newblock Diffusart: Enhancing line art colorization with conditional diffusion models.
\newblock In \emph{Proceedings of the IEEE/CVF Conference on Computer Vision and Pattern Recognition}, pages 3486--3490, 2023.

\bibitem[Chan et~al.(2022)Chan, Durand, and Isola]{chan2022learning}
Caroline Chan, Fr{\'e}do Durand, and Phillip Isola.
\newblock Learning to generate line drawings that convey geometry and semantics.
\newblock In \emph{Proceedings of the IEEE/CVF Conference on Computer Vision and Pattern Recognition}, pages 7915--7925, 2022.

\bibitem[Chen and Contributors(2024)]{animegan}
Lei Chen and Contributors.
\newblock Animegan: A fast and simple image animation method, 2024.

\bibitem[Chen and Zwicker(2022)]{chen2022eisai}
Shuhong Chen and Matthias Zwicker.
\newblock Improving the perceptual quality of 2d animation interpolation.
\newblock In \emph{European Conference on Computer Vision}, pages 271--287. Springer, 2022.

\bibitem[Chen et~al.(2020)Chen, Zhang, Gao, He, Xia, Shi, and Zhang]{chen2020active}
Shu-Yu Chen, Jia-Qi Zhang, Lin Gao, Yue He, Shihong Xia, Min Shi, and Fang-Lue Zhang.
\newblock Active colorization for cartoon line drawings.
\newblock \emph{IEEE Transactions on Visualization and Computer Graphics}, 28\penalty0 (2):\penalty0 1198--1208, 2020.

\bibitem[Chen et~al.(2023)Chen, Wang, Zhang, Zhuang, Ma, Yu, Wang, Lin, Qiao, and Liu]{chen2023seine}
Xinyuan Chen, Yaohui Wang, Lingjun Zhang, Shaobin Zhuang, Xin Ma, Jiashuo Yu, Yali Wang, Dahua Lin, Yu Qiao, and Ziwei Liu.
\newblock Seine: Short-to-long video diffusion model for generative transition and prediction.
\newblock In \emph{The Twelfth International Conference on Learning Representations}, 2023.

\bibitem[Ci et~al.(2018)Ci, Ma, Wang, Li, and Luo]{Ci_2018}
Yuanzheng Ci, Xinzhu Ma, Zhihui Wang, Haojie Li, and Zhongxuan Luo.
\newblock User-guided deep anime line art colorization with conditional adversarial networks.
\newblock In \emph{Proceedings of the 26th ACM international conference on Multimedia}, 2018.

\bibitem[Danier et~al.(2024)Danier, Zhang, and Bull]{danier2024ldmvfi}
Duolikun Danier, Fan Zhang, and David Bull.
\newblock {LDMVFI:} video frame interpolation with latent diffusion models.
\newblock In \emph{Assoc. Adv. Artif. Intell.}, 2024.

\bibitem[Dou et~al.(2021)Dou, Wang, Li, Wang, Li, and Liu]{dou2021dual}
Zhi Dou, Ning Wang, Baopu Li, Zhihui Wang, Haojie Li, and Bin Liu.
\newblock Dual color space guided sketch colorization.
\newblock \emph{IEEE Trans. Image Process.}, 30:\penalty0 7292--7304, 2021.

\bibitem[Fourey et~al.(2018)Fourey, Tschumperlé, and Revoy]{Fourey_Tschumperlé_Revoy_2018}
Sébastien Fourey, David Tschumperlé, and David Revoy.
\newblock A fast and efficient semi-guided algorithm for flat coloring line-arts.
\newblock \emph{Le Centre pour la Communication Scientifique Directe - HAL - Archive ouverte HAL,Le Centre pour la Communication Scientifique Directe - HAL - Archive ouverte HAL}, 2018.

\bibitem[Furusawa et~al.(2017)Furusawa, Hiroshiba, Ogaki, and Odagiri]{Furusawa_Hiroshiba_Ogaki_Odagiri_2017}
Chie Furusawa, Kazuyuki Hiroshiba, Keisuke Ogaki, and Yuri Odagiri.
\newblock Comicolorization: Semi-automatic manga colorization.
\newblock \emph{Cornell University - arXiv,Cornell University - arXiv}, 2017.

\bibitem[Guo et~al.(2024)Guo, Yang, Rao, Agrawala, Lin, and Dai]{Guo_2023}
Yuwei Guo, Ceyuan Yang, Anyi Rao, Maneesh Agrawala, Dahua Lin, and Bo Dai.
\newblock Sparsectrl: Adding sparse controls to text-to-video diffusion models.
\newblock In \emph{Eur. Conf. Comput. Vis.}, pages 330--348, 2024.

\bibitem[He et~al.(2024)He, Liu, Qian, Wang, Hu, Cao, Yan, and Zhang]{he2024id}
Xuanhua He, Quande Liu, Shengju Qian, Xin Wang, Tao Hu, Ke Cao, Keyu Yan, and Jie Zhang.
\newblock Id-animator: Zero-shot identity-preserving human video generation.
\newblock \emph{arXiv preprint arXiv:2404.15275}, 2024.

\bibitem[Heusel et~al.(2017)Heusel, Ramsauer, Unterthiner, Nessler, and Hochreiter]{heusel2017fid}
Martin Heusel, Hubert Ramsauer, Thomas Unterthiner, Bernhard Nessler, and Sepp Hochreiter.
\newblock Gans trained by a two time-scale update rule converge to a local nash equilibrium.
\newblock \emph{Advances in neural information processing systems}, 30, 2017.

\bibitem[Huang et~al.(2020)Huang, Zhang, Heng, Shi, and Zhou]{huang2020rife}
Zhewei Huang, Tianyuan Zhang, Wen Heng, Boxin Shi, and Shuchang Zhou.
\newblock {RIFE:} real-time intermediate flow estimation for video frame interpolation.
\newblock \emph{CoRR}, abs/2011.06294, 2020.

\bibitem[Huang et~al.(2022)Huang, Zhao, and Liao]{huang2022unicolor}
Zhitong Huang, Nanxuan Zhao, and Jing Liao.
\newblock Unicolor: A unified framework for multi-modal colorization with transformer.
\newblock \emph{ACM Transactions on Graphics (TOG)}, 41\penalty0 (6):\penalty0 1--16, 2022.

\bibitem[Huang et~al.(2024)Huang, Zhang, and Liao]{huang2024lvcd}
Zhitong Huang, Mohan Zhang, and Jing Liao.
\newblock Lvcd: reference-based lineart video colorization with diffusion models.
\newblock \emph{arXiv preprint arXiv:2409.12960}, 2024.

\bibitem[Jain et~al.(2024)Jain, Watson, Tabellion, Holynski, Poole, and Kontkanen]{jain2024video}
Siddhant Jain, Daniel Watson, Eric Tabellion, Aleksander Holynski, Ben Poole, and Janne Kontkanen.
\newblock Video interpolation with diffusion models.
\newblock \emph{CoRR}, abs/2404.01203, 2024.

\bibitem[Jiang et~al.(2018)Jiang, Sun, Jampani, Yang, Learned{-}Miller, and Kautz]{superslomo2018}
Huaizu Jiang, Deqing Sun, Varun Jampani, Ming{-}Hsuan Yang, Erik~G. Learned{-}Miller, and Jan Kautz.
\newblock Super slomo: High quality estimation of multiple intermediate frames for video interpolation.
\newblock In \emph{IEEE Conf. Comput. Vis. Pattern Recog.}, 2018.

\bibitem[Karaev et~al.(2023)Karaev, Rocco, Graham, Neverova, Vedaldi, and Rupprecht]{karaev2023cotracker}
Nikita Karaev, Ignacio Rocco, Benjamin Graham, Natalia Neverova, Andrea Vedaldi, and Christian Rupprecht.
\newblock Cotracker: It is better to track together.
\newblock \emph{arXiv preprint arXiv:2307.07635}, 2023.

\bibitem[Kim et~al.(2019{\natexlab{a}})Kim, Jhoo, Park, and Yoo]{kim2019tag2pix}
Hyunsu Kim, Ho~Young Jhoo, Eunhyeok Park, and Sungjoo Yoo.
\newblock Tag2pix: Line art colorization using text tag with secat and changing loss.
\newblock In \emph{Int. Conf. Comput. Vis.}, pages 9056--9065, 2019{\natexlab{a}}.

\bibitem[Kim et~al.(2019{\natexlab{b}})Kim, Jhoo, Park, and Yoo]{Kim_2019}
Hyun-Su Kim, HoYoung Jhoo, Eunhyeok Park, and Sungjoo Yoo.
\newblock Tag2pix: Line art colorization using text tag with secat and changing loss.
\newblock \emph{Cornell University - arXiv,Cornell University - arXiv}, 2019{\natexlab{b}}.

\bibitem[Li et~al.(2022{\natexlab{a}})Li, Wu, Sun, Tao, Tang, and Tai]{hvfi}
Changlin Li, Guangyang Wu, Yanan Sun, Xin Tao, Chi{-}Keung Tang, and Yu{-}Wing Tai.
\newblock {H-VFI:} hierarchical frame interpolation for videos with large motions.
\newblock \emph{CoRR}, abs/2211.11309, 2022{\natexlab{a}}.

\bibitem[Li et~al.(2022{\natexlab{b}})Li, Geng, Kang, Chen, and Yang]{li2022eliminating}
Zekun Li, Zhengyang Geng, Zhao Kang, Wenyu Chen, and Yibo Yang.
\newblock Eliminating gradient conflict in reference-based line-art colorization.
\newblock In \emph{European Conference on Computer Vision}, pages 579--596. Springer, 2022{\natexlab{b}}.

\bibitem[Lindenberger et~al.(2023)Lindenberger, Sarlin, and Pollefeys]{lightglue}
Philipp Lindenberger, Paul-Edouard Sarlin, and Marc Pollefeys.
\newblock Lightglue: Local feature matching at light speed.
\newblock In \emph{Proceedings of the IEEE/CVF International Conference on Computer Vision}, pages 17627--17638, 2023.

\bibitem[Liu et~al.(2018)Liu, Qin, Wan, and Luo]{liu2018auto}
Yifan Liu, Zengchang Qin, Tao Wan, and Zhenbo Luo.
\newblock Auto-painter: Cartoon image generation from sketch by using conditional wasserstein generative adversarial networks.
\newblock \emph{Neurocomputing}, 311:\penalty0 78--87, 2018.

\bibitem[Lowe(2004)]{lowe2004sift}
David~G Lowe.
\newblock Distinctive image features from scale-invariant keypoints.
\newblock \emph{International journal of computer vision}, 60:\penalty0 91--110, 2004.

\bibitem[Maejima et~al.(2019)Maejima, Kubo, Funatomi, Yotsukura, Nakamura, and Mukaigawa]{maejima2019graph}
Akinobu Maejima, Hiroyuki Kubo, Takuya Funatomi, Tatsuo Yotsukura, Satoshi Nakamura, and Yasuhiro Mukaigawa.
\newblock Graph matching based anime colorization with multiple references.
\newblock In \emph{Special Interest Group on Computer Graphics and Interactive Techniques Conference, {SIGGRAPH} 2019, Los Angeles, CA, USA, July 28 - August 1, 2019, Posters}, pages 13:1--13:2, 2019.

\bibitem[Pan et~al.(2024)Pan, Zhu, and Mu]{pan2024sakuga}
Zhenglin Pan, Yu Zhu, and Yuxuan Mu.
\newblock Sakuga-42m dataset: Scaling up cartoon research.
\newblock \emph{arXiv preprint arXiv:2405.07425}, 2024.

\bibitem[Parakkat et~al.(2022)Parakkat, Memari, and Cani]{Parakkat_Memari_Cani_2022}
Amal~Dev Parakkat, Pooran Memari, and Marie‐Paule Cani.
\newblock Delaunay painting: Perceptual image colouring from raster contours with gaps.
\newblock \emph{Computer Graphics Forum}, page 166–181, 2022.

\bibitem[Qin et~al.(2020)Qin, Zhang, Huang, Dehghan, Zaiane, and Jagersand]{u2net}
Xuebin Qin, Zichen Zhang, Chenyang Huang, Masood Dehghan, Osmar~R Zaiane, and Martin Jagersand.
\newblock U2-net: Going deeper with nested u-structure for salient object detection.
\newblock \emph{Pattern recognition}, 106:\penalty0 107404, 2020.

\bibitem[Qu et~al.(2006)Qu, Wong, and Heng]{Qu_2006}
Yingge Qu, Tien-Tsin Wong, and Pheng-Ann Heng.
\newblock Manga colorization.
\newblock \emph{ACM Transactions on Graphics}, 25\penalty0 (3):\penalty0 1214–1220, 2006.

\bibitem[Rombach et~al.(2022)Rombach, Blattmann, Lorenz, Esser, and Ommer]{sd}
Robin Rombach, Andreas Blattmann, Dominik Lorenz, Patrick Esser, and Bj{\"o}rn Ommer.
\newblock High-resolution image synthesis with latent diffusion models.
\newblock In \emph{Proceedings of the IEEE/CVF conference on computer vision and pattern recognition}, pages 10684--10695, 2022.

\bibitem[Sangkloy et~al.(2017)Sangkloy, Lu, Fang, Yu, and Hays]{sangkloy2017scribbler}
Patsorn Sangkloy, Jingwan Lu, Chen Fang, Fisher Yu, and James Hays.
\newblock Scribbler: Controlling deep image synthesis with sketch and color.
\newblock In \emph{IEEE Conf. Comput. Vis. Pattern Recog.}, pages 5400--5409, 2017.

\bibitem[Sato et~al.(2014)Sato, Matsui, Yamasaki, and Aizawa]{Sato_2014}
Kazuhiro Sato, Yusuke Matsui, Toshihiko Yamasaki, and Kiyoharu Aizawa.
\newblock Reference-based manga colorization by graph correspondence using quadratic programming.
\newblock In \emph{SIGGRAPH Asia 2014 Technical Briefs}, page 1–4, 2014.

\bibitem[Shi et~al.(2020)Shi, Zhang, Chen, Gao, Lai, and Zhang]{shi2020deep}
Min Shi, Jia-Qi Zhang, Shu-Yu Chen, Lin Gao, Yu-Kun Lai, and Fang-Lue Zhang.
\newblock Deep line art video colorization with a few references.
\newblock \emph{arXiv preprint arXiv:2003.10685}, 2020.

\bibitem[Siyao et~al.(2021)Siyao, Zhao, Yu, Sun, Metaxas, Loy, and Liu]{siyao2021AnimeInterp}
Li Siyao, Shiyu Zhao, Weijiang Yu, Wenxiu Sun, Dimitris Metaxas, Chen~Change Loy, and Ziwei Liu.
\newblock Deep animation video interpolation in the wild.
\newblock In \emph{Proceedings of the IEEE/CVF conference on computer vision and pattern recognition}, pages 6587--6595, 2021.

\bibitem[Su et~al.(2021)Su, Liu, Yu, Hu, Liao, Tian, Pietik{\"a}inen, and Liu]{su2021pidinet}
Zhuo Su, Wenzhe Liu, Zitong Yu, Dewen Hu, Qing Liao, Qi Tian, Matti Pietik{\"a}inen, and Li Liu.
\newblock Pixel difference networks for efficient edge detection.
\newblock In \emph{Proceedings of the IEEE/CVF international conference on computer vision}, pages 5117--5127, 2021.

\bibitem[Sýkora et~al.(2009{\natexlab{a}})Sýkora, Dingliana, and Collins]{S_2009}
Daniel Sýkora, John Dingliana, and Steven Collins.
\newblock Lazybrush: Flexible painting tool for hand-drawn cartoons.
\newblock \emph{Computer Graphics Forum}, 28\penalty0 (2):\penalty0 599–608, 2009{\natexlab{a}}.

\bibitem[Sýkora et~al.(2009{\natexlab{b}})Sýkora, Dingliana, and Collins]{Sýkora_Dingliana_Collins_2009}
Daniel Sýkora, John Dingliana, and Steven Collins.
\newblock Lazybrush: Flexible painting tool for hand-drawn cartoons.
\newblock \emph{Computer Graphics Forum}, page 599–608, 2009{\natexlab{b}}.

\bibitem[Tang et~al.(2023)Tang, Jia, Wang, Phoo, and Hariharan]{dift}
Luming Tang, Menglin Jia, Qianqian Wang, Cheng~Perng Phoo, and Bharath Hariharan.
\newblock Emergent correspondence from image diffusion.
\newblock \emph{Advances in Neural Information Processing Systems}, 36:\penalty0 1363--1389, 2023.

\bibitem[Thasarathan et~al.(2019)Thasarathan, Nazeri, and Ebrahimi]{tcvc}
Harrish Thasarathan, Kamyar Nazeri, and Mehran Ebrahimi.
\newblock Automatic temporally coherent video colorization.
\newblock In \emph{2019 16th conference on computer and robot vision (CRV)}, pages 189--194. IEEE, 2019.

\bibitem[Unterthiner et~al.(2018)Unterthiner, Van~Steenkiste, Kurach, Marinier, Michalski, and Gelly]{unterthiner2018fvd}
Thomas Unterthiner, Sjoerd Van~Steenkiste, Karol Kurach, Raphael Marinier, Marcin Michalski, and Sylvain Gelly.
\newblock Towards accurate generative models of video: A new metric \& challenges.
\newblock \emph{arXiv preprint arXiv:1812.01717}, 2018.

\bibitem[Wang et~al.(2023)Wang, Niu, Dou, Wang, Wang, Ming, Liu, and Li]{wang2023TRE-Net}
Ning Wang, Muyao Niu, Zhi Dou, Zhihui Wang, Zhiyong Wang, Zhaoyan Ming, Bin Liu, and Haojie Li.
\newblock Coloring anime line art videos with transformation region enhancement network.
\newblock \emph{Pattern Recognition}, 141:\penalty0 109562, 2023.

\bibitem[Wang et~al.(2024)Wang, Wang, Zheng, Ouyang, Chen, Gong, Chen, Shen, and Shen]{wang2024framer}
Wen Wang, Qiuyu Wang, Kecheng Zheng, Hao Ouyang, Zhekai Chen, Biao Gong, Hao Chen, Yujun Shen, and Chunhua Shen.
\newblock Framer: Interactive video interpolation.
\newblock \emph{arXiv preprint https://arxiv.org/abs/2410.18978}, 2024.

\bibitem[Wu et~al.(2025)Wu, Li, Gu, Zhao, He, Zhang, Shou, Li, Gao, and Zhang]{wu2025draganything}
Weijia Wu, Zhuang Li, Yuchao Gu, Rui Zhao, Yefei He, David~Junhao Zhang, Mike~Zheng Shou, Yan Li, Tingting Gao, and Di Zhang.
\newblock Draganything: Motion control for anything using entity representation.
\newblock In \emph{European Conference on Computer Vision}, pages 331--348. Springer, 2025.

\bibitem[Xie and Tu(2015)]{xie2015hed}
Saining Xie and Zhuowen Tu.
\newblock Holistically-nested edge detection.
\newblock In \emph{Proceedings of the IEEE international conference on computer vision}, pages 1395--1403, 2015.

\bibitem[Xing et~al.(2024)Xing, Liu, Xia, Zhang, Wang, Shan, and Wong]{xing2024tooncrafter}
Jinbo Xing, Hanyuan Liu, Menghan Xia, Yong Zhang, Xintao Wang, Ying Shan, and Tien-Tsin Wong.
\newblock Tooncrafter: Generative cartoon interpolation.
\newblock \emph{arXiv preprint arXiv:2405.17933}, 2024.

\bibitem[Yan et~al.(2024)Yan, Yuan, Wu, Nishioka, Fujishiro, and Saito]{yan2024colorizediffusion}
Dingkun Yan, Liang Yuan, Erwin Wu, Yuma Nishioka, Issei Fujishiro, and Suguru Saito.
\newblock Colorizediffusion: Adjustable sketch colorization with reference image and text.
\newblock \emph{arXiv preprint arXiv:2401.01456}, 2024.

\bibitem[Yang et~al.(2025)Yang, Zeng, Zhang, and Zhang]{x-pose}
Jie Yang, Ailing Zeng, Ruimao Zhang, and Lei Zhang.
\newblock X-pose: Detecting any keypoints.
\newblock In \emph{European Conference on Computer Vision}, pages 249--268. Springer, 2025.

\bibitem[Ye et~al.(2023)Ye, Zhang, Liu, Han, and Yang]{ye2023ip}
Hu Ye, Jun Zhang, Sibo Liu, Xiao Han, and Wei Yang.
\newblock Ip-adapter: Text compatible image prompt adapter for text-to-image diffusion models.
\newblock \emph{arXiv preprint arXiv:2308.06721}, 2023.

\bibitem[Yu et~al.(2024)Yu, Qian, Wang, Dong, and Liu]{yu2024ACOF}
Yifeng Yu, Jiangbo Qian, Chong Wang, Yihong Dong, and Baisong Liu.
\newblock Animation line art colorization based on the optical flow method.
\newblock \emph{Computer Animation and Virtual Worlds}, 35\penalty0 (1):\penalty0 e2229, 2024.

\bibitem[Zhang et~al.(2018{\natexlab{a}})Zhang, Li, Wong, Ji, and Liu]{Zhang_2018}
Lvmin Zhang, Chengze Li, Tien-Tsin Wong, Yi Ji, and Chunping Liu.
\newblock Two-stage sketch colorization.
\newblock \emph{ACM Transactions on Graphics}, page 1–14, 2018{\natexlab{a}}.

\bibitem[Zhang et~al.(2018{\natexlab{b}})Zhang, Li, Wong, Ji, and Liu]{zhang2018two}
Lvmin Zhang, Chengze Li, Tien-Tsin Wong, Yi Ji, and Chunping Liu.
\newblock Two-stage sketch colorization.
\newblock \emph{ACM Trans. Graph.}, 37\penalty0 (6):\penalty0 1--14, 2018{\natexlab{b}}.

\bibitem[Zhang et~al.(2021{\natexlab{a}})Zhang, Li, Simo-Serra, Ji, Wong, and Liu]{zhang2021user}
Lvmin Zhang, Chengze Li, Edgar Simo-Serra, Yi Ji, Tien-Tsin Wong, and Chunping Liu.
\newblock User-guided line art flat filling with split filling mechanism.
\newblock In \emph{Proceedings of the IEEE/CVF conference on computer vision and pattern recognition}, pages 9889--9898, 2021{\natexlab{a}}.

\bibitem[Zhang et~al.(2023{\natexlab{a}})Zhang, Rao, and Agrawala]{zhang2023adding}
Lvmin Zhang, Anyi Rao, and Maneesh Agrawala.
\newblock Adding conditional control to text-to-image diffusion models.
\newblock In \emph{Int. Conf. Comput. Vis.}, pages 3836--3847, 2023{\natexlab{a}}.

\bibitem[Zhang et~al.(2023{\natexlab{b}})Zhang, Rao, and Agrawala]{zhang2023controlnet}
Lvmin Zhang, Anyi Rao, and Maneesh Agrawala.
\newblock Adding conditional control to text-to-image diffusion models.
\newblock In \emph{Proceedings of the IEEE/CVF International Conference on Computer Vision}, pages 3836--3847, 2023{\natexlab{b}}.

\bibitem[Zhang et~al.(2020)Zhang, Bo, Wang, Li, and Hui]{Zhang_2020}
Qian Zhang, Wang Bo, Wen Wang, Hai Li, and Liu Hui.
\newblock Line art correlation matching feature transfer network for automatic animation colorization.
\newblock \emph{Cornell University - arXiv,Cornell University - arXiv}, 2020.

\bibitem[Zhang et~al.(2021{\natexlab{b}})Zhang, Wang, Wen, Li, and Liu]{zhang2021line}
Qian Zhang, Bo Wang, Wei Wen, Hai Li, and Junhui Liu.
\newblock Line art correlation matching feature transfer network for automatic animation colorization.
\newblock In \emph{Proceedings of the IEEE/CVF Winter Conference on Applications of Computer Vision}, pages 3872--3881, 2021{\natexlab{b}}.

\bibitem[Zhou et~al.(2021)Zhou, Zhang, Zhang, Zhang, Bao, Chen, Zhang, and Wen]{zhou2021cocosnet}
Xingran Zhou, Bo Zhang, Ting Zhang, Pan Zhang, Jianmin Bao, Dong Chen, Zhongfei Zhang, and Fang Wen.
\newblock Cocosnet v2: Full-resolution correspondence learning for image translation.
\newblock In \emph{Proceedings of the IEEE/CVF conference on computer vision and pattern recognition}, pages 11465--11475, 2021.

\bibitem[Zhu et~al.(2024)Zhu, Shang, Ren, and Zuo]{zhu2024thin}
Tianyi Zhu, Wei Shang, Dongwei Ren, and Wangmeng Zuo.
\newblock Thin-plate spline-based interpolation for animation line inbetweening.
\newblock \emph{arXiv preprint arXiv:2408.09131}, 2024.

\bibitem[Zou et~al.(2019)Zou, Mo, Gao, Du, and Fu]{Zou_2019}
Changqing Zou, Haoran Mo, Chengying Gao, Ruofei Du, and Hongbo Fu.
\newblock Language-based colorization of scene sketches.
\newblock \emph{ACM Transactions on Graphics}, page 1–16, 2019.

\bibitem[Zou et~al.(2024)Zou, Wan, Blanch, Murn, Mrak, Sock, Yang, and Herranz]{zou2024lightweight}
Chengyi Zou, Shuai Wan, Marc~Gorriz Blanch, Luka Murn, Marta Mrak, Juil Sock, Fei Yang, and Luis Herranz.
\newblock Lightweight deep exemplar colorization via semantic attention-guided laplacian pyramid.
\newblock \emph{IEEE Transactions on Visualization and Computer Graphics}, 2024.

\end{thebibliography}
}

\clearpage
\appendix
\renewcommand\thesection{\Alph{section}}
\renewcommand\thefigure{S\arabic{figure}}
\renewcommand\thetable{S\arabic{table}}
\renewcommand\theequation{S\arabic{equation}}
\setcounter{figure}{0}
\setcounter{table}{0}
\setcounter{equation}{0}
\setcounter{page}{1}
\maketitlesupplementary

\section*{Appendix}

\section{Reference with Different Background}\label{sec:different_background}
\begin{figure}[ht]
    \centering
    \includegraphics[width=1.0\linewidth]{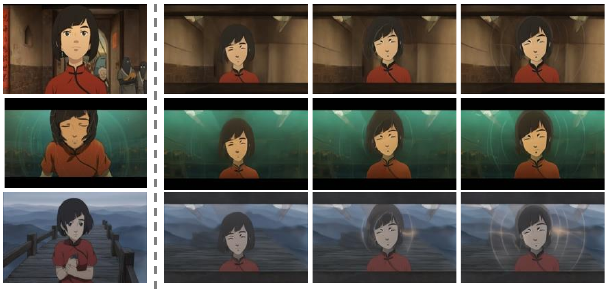}
    \caption{Illustration of reference with different backgrounds. }
    \label{fig:diffbg}

\end{figure}

When using images of the same character with different backgrounds as references, our model can transfer the style from the reference images to generate new backgrounds with diverse styles. The original character remains consistent in their core features, such as expressions and clothing, while the integration of varied backgrounds enriches the overall visual effect, as shown in~\cref{fig:diffbg}.

\section{Multiple Characters}
\begin{figure}[ht]
    \centering
    \includegraphics[width=1.0\linewidth]{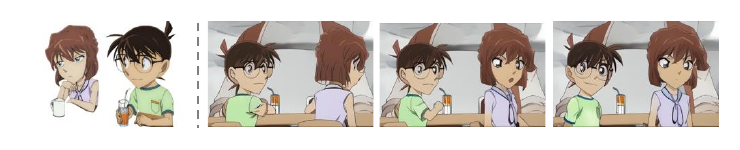}
    \caption{Illustration of the multiple characters situation. When the reference image contains multiple characters, our method can correctly infer the correspondence and apply colorization to each character accordingly. }
    \label{fig:multiple}
    \vspace{-10pt}
\end{figure}

Although our work primarily focuses on a single reference image and does not include specific training or processing for handling multiple references, we observe that our model can automatically distinguish between multiple characters in a reference image based on their unique features. It applies the correct coloring to each character, even when there are significant differences in poses, angles, or relative positions between the reference and the line art, as demonstrated in~\cref{fig:multiple}.

\section{Different Line Art Extraction Methods}

To evaluate the generalization capability of our method under different line art conditions, we test its performance using various line art extraction methods. Besides the default line art extraction method~\cite{chan2022learning} used in our paper, we also apply three additional methods: Anime Lineart~\cite{animegan}, HED~\cite{xie2015hed} and PiDiNet~\cite{su2021pidinet}. Among these, Anime Lineart is a line art extraction method specifically trained on anime datasets. HED, as an edge detection method, produces relatively thick line art, whereas PiDiNet creates simplistic line art that is closer to hand-drawn style.

After extraction, we apply the same binarization process to the line art as described in the main text. Our method successfully colors the line art under different conditions while maintaining consistency with the reference. Due to the varying characteristics of the extracted line art, our method generates different results accordingly. 

\begin{figure}[ht]
    \centering
    \includegraphics[width=1.0\linewidth]{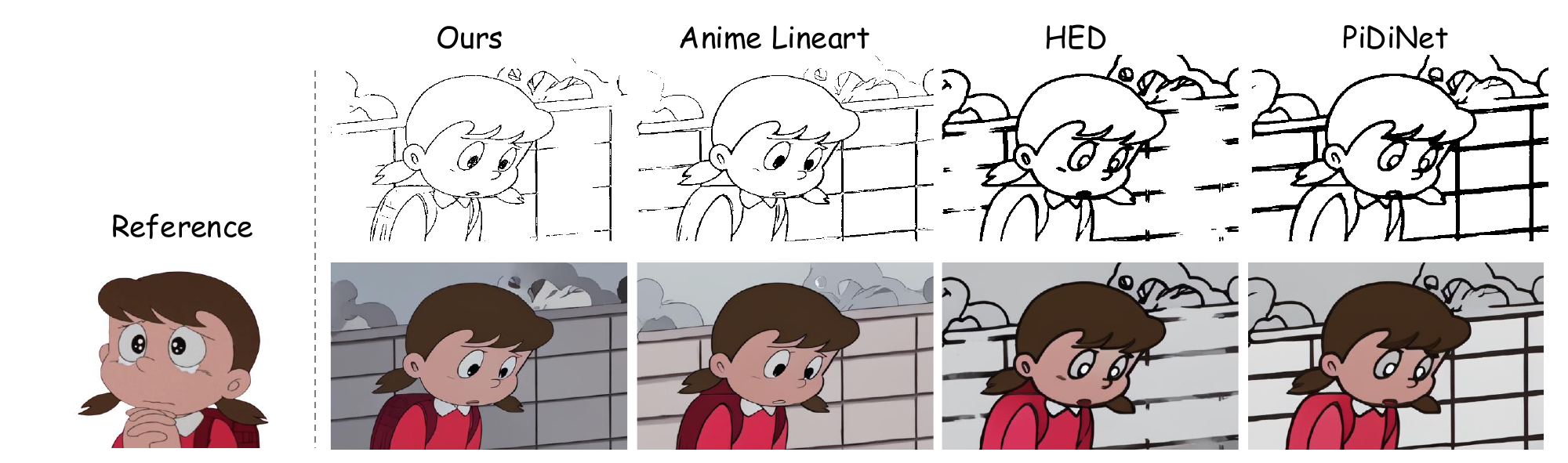}
    \caption{Impact of different line art extraction methods.}
    \label{fig:diff_sketch}
   
\end{figure}
\section{Motivation for Correspondence Matching}\label{sec:correspondence}

\begin{figure}[ht]
    \centering
    \includegraphics[width=1.0\linewidth]{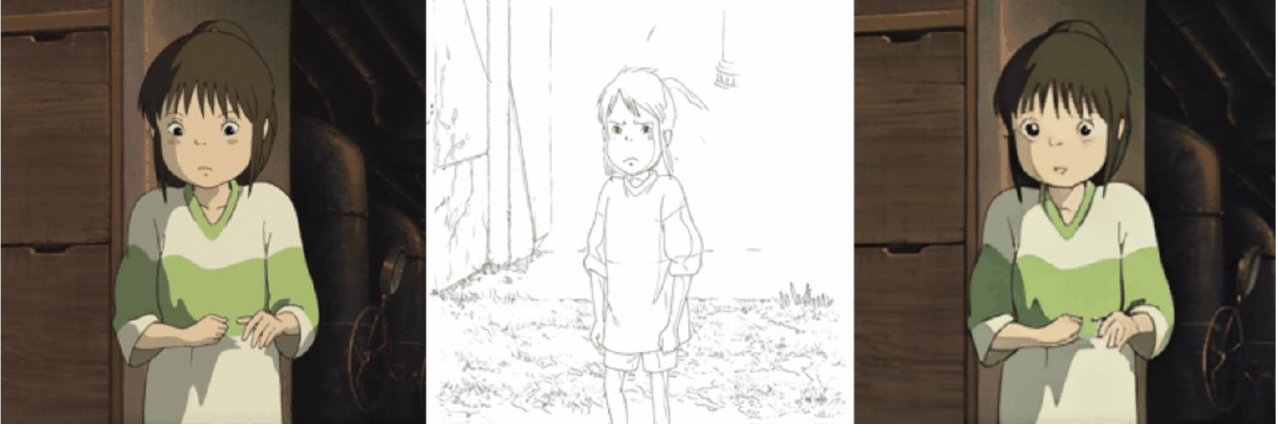}
    \caption{In the early training stage (10k step), the video generation model produces static videos that closely resemble the given reference design.}
    \label{fig:svd_prior}
\end{figure}

As an image-to-video model, SVD (Stable Video Diffusion)~\cite{svd} inherently possesses the ability to extract information from an input image to generate a video. However, during training, we observe that the strong prior in SVD restricts the first frame to be the same with the input reference image, as shown in~\cref{fig:svd_prior}.

In our formulation, the input image is not the first frame of the video but rather a reference character design from a different viewpoint. The model needs to query colors from this reference image, while the structure information should align with the given sketch list. This conflicting prior makes training the model significantly more challenging.

To better establish the relationship between the reference character design and the sketch, reduce the learning difficulty for the model, and improve the fine-grained details, we propose a Correspondence Matching Module. This module explicitly injects the matching relationships between the reference image and the sketch, enabling the model to better query and color the correct areas.

\section{Illustration of DIFT Matching}\label{sec:dift}

\begin{figure}[ht]
    \centering
    \includegraphics[width=1.0\linewidth]{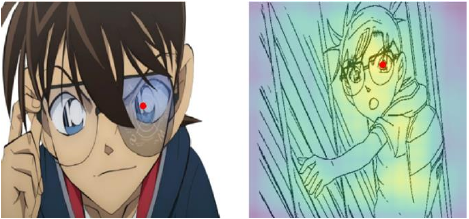}
    \caption{Semantic feature can effectively find matching keypoints between reference color image and binarized sketch.  }
    \label{fig:dift_matching}
\end{figure}

During training, we apply low-level techniques LightGlue~\cite{lightglue} with SIFT descriptor~\cite{lowe2004sift} for keypoint selection and matching between the reference image and the training video frames due to its fast speed. During inference, we lack access to the ground truth color image. Techniques that rely on low-level image features, such as SIFT descriptors, are ineffective at accurately matching keypoints between sketches and color reference images due to the significant domain gap between them. Therefore, we use the semantic level keypoint matching method DIFT~\cite{dift} to establish the correspondence between the color reference image and the sketches, as shown in~\cref{fig:dift_matching}.

\section{Limitation}\label{sec:limitation}

Although our method can colorize multiple clips containing the same character based on a single character design sheet while maintaining good character consistency, it still has certain limitations.

\begin{figure}[ht]
    \centering
    \includegraphics[width=1.0\linewidth]{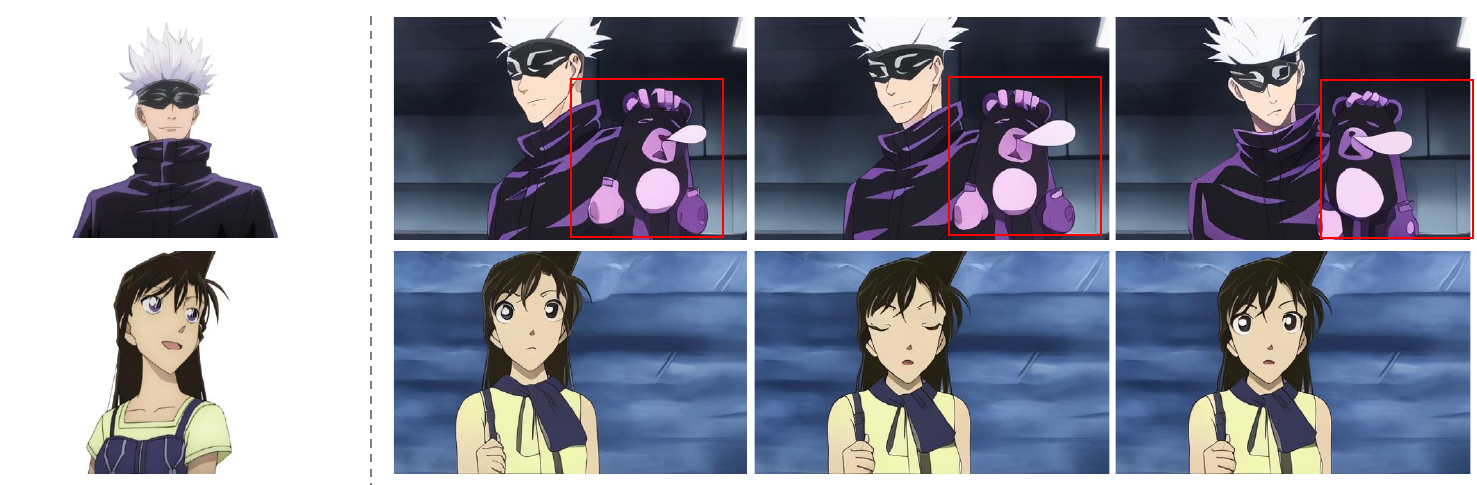}
    \caption{Limitations: in the $1^{st}$ row, the cartoon bear highlighted within the red square is not present in the reference image. Consequently, the model can only infer the bear’s color as purple, based on the main color of the reference character, which deviates from its actual appearance. In the $2^{nd}$ row, the character's clothing in the line art clip is different from the reference. Therefore, our model can only infer the color of the dress and scarf based on the dominant color patterns identified in the reference image.}
    \label{fig:failure}
   
\end{figure}

First, when a line art clip contains objects that are not present in the reference, the model struggles to determine the appropriate colors for these objects, as shown in the $1^{st}$ row of~\cref{fig:failure}. It can only infer colors based on the color information available in the reference, leading to inaccuracies in the colorization.

Second, when the clothing of a character in the line art clip differs from that in the reference image (even though it is the same character), our model can only infer reasonable colors based on the color patterns of the character's clothing in the reference image. However, our method cannot guarantee accuracy in this situation, as shown in the $2^{nd}$ row of~\cref{fig:failure}.

%   

% {
% % \clearpage
% \small
% \bibliographystyle{ieeenat_fullname}
% \bibliography{ref.bib}
% }

\end{document}